\documentclass[conference]{IEEEtran}

\usepackage[T1]{fontenc}
\usepackage[utf8]{inputenc}
\usepackage{amsmath}
\usepackage{amssymb}
\usepackage{graphicx}
\usepackage{booktabs}
\usepackage{array}
\usepackage{algorithm}
\usepackage{algpseudocode}
\usepackage{listings}
\usepackage{cite}
\usepackage{url}
\usepackage[hidelinks]{hyperref}

\algrenewcommand\algorithmicrequire{\textbf{Input:}}
\algrenewcommand\algorithmicensure{\textbf{Output:}}
\algrenewcommand\algorithmiccomment[1]{\hskip1.5em$\triangleright$\ #1}

\lstdefinelanguage{Cypher}{
    morekeywords={CALL, UNWIND, WITH, WHERE, MATCH, OPTIONAL, RETURN, AS, AND, OR,
        IS, NOT, NULL, CASE, WHEN, THEN, ELSE, END, IN, UNION, ORDER, BY, DESC, LIMIT},
    morekeywords=[2]{collect, sum, max, size, nodes, keys, coalesce, reduce, range,
        gds, apoc, db},
    sensitive=true,
    morecomment=[l]{//},
    morestring=[b]",
    morestring=[b]'
}
\makeatletter
\AtBeginDocument{%
  \long\def\lst@makecaption#1#2{%
    \def\@captype{lstlisting}%
    \@makecaption{#1}{#2}%
    \par\nobreak\vskip\belowcaptionskip}%
}
\makeatother

\graphicspath{{../computations/}}

\begin{document}

\title{Query-Aware Spreading Activation for Multi-Hop Retrieval over Knowledge Graphs}

\author{%
    \IEEEauthorblockN{Mykola M. Glybovets}
    \IEEEauthorblockA{%
        \textit{National University of Kyiv-Mohyla Academy}\\
        Kyiv, Ukraine\\
        glib@ukma.edu.ua}
    \and
    \IEEEauthorblockN{Illia S. Makarov}
    \IEEEauthorblockA{%
        \textit{National University of Kyiv-Mohyla Academy}\\
        Kyiv, Ukraine\\
        i.makarov@ukma.edu.ua}
}

\maketitle

\begin{abstract}
Retrieval-augmented generation built on knowledge graphs (Graph RAG) outperforms flat passage retrieval on multi-hop question answering by leveraging graph structure. In most existing systems, however, the question only sets the seed nodes; the subsequent traversal becomes ``query-blind'', depending solely on the graph structure. The exception is QAFD-RAG, which implements query-aware traversal via a flow-diffusion solver with combined edge re-weighting. This architecture requires loading the full graph into Python memory and an iterative solver with a variable number of iterations complicating integration with the graph database. We propose a spreading-activation method that achieves the same query-aware traversal with a single per-step semantic gate: the step weight is the cosine similarity between the candidate entity's description and the question, and the number of iterations is fixed. The whole retrieval procedure --- seed mapping, propagation, top-$K$ selection and context assembly --- is expressed as a single Cypher query executed in one round-trip to Neo4j; the graph never leaves the database. On MuSiQue our method matches QAFD-RAG by exact match (32.80 vs 33.50) and outperforms the strongest purely-structural baseline in our comparison, HippoRAG, by 5.3 EM and 3.4 $F_1$; on 2WikiMultiHopQA HippoRAG and QAFD-RAG retain an advantage due to their phrase-node architectures. An ablation with the gate disabled confirms that the gate is the source of a simultaneous $F_1$ gain of 3.6 to 7.4 points and a retrieval-latency reduction by a factor of 1.5 to 4.9.
\end{abstract}



\section{Introduction}

Retrieval-augmented generation (RAG) supplements the parametric knowledge of large 
language models with external evidence incorporated directly at 
inference time~\cite{lewis2020rag, gao2023survey}. The dominant retrieval primitive 
is dense passage retrieval~\cite{karpukhin2020dpr}: the corpus is segmented into passages, 
each passage is embedded into a vector space, and the $k$ passages most similar to the target question 
are added to the language model's context. This scheme is highly effective for questions 
whose answer is contained within a single passage, but its quality degrades sharply on 
multi-hop questions that require synthesizing evidence from several documents, 
which may share no surface lexical overlap. This limitation is structural in nature: 
the question and the answer-bearing passage are separated by a chain of inference that vector 
similarity cannot reconstruct.

A growing body of work overcomes this limitation by indexing the corpus as a knowledge graph 
and leveraging its structure during retrieval~\cite{edge2024graphrag, guo2024lightrag, gutierrez2024hipporag, gutierrez2025hipporag2, chen2025pathrag}. 
Despite differences in the traversal algorithm --- community detection (GraphRAG~\cite{edge2024graphrag}), 
personalized PageRank (HippoRAG~\cite{gutierrez2024hipporag}), 
flow-based path pruning (PathRAG~\cite{chen2025pathrag}) --- these systems are alike in one respect: 
the question influences only the selection of seed nodes. 
The traversal itself depends only on the graph structure and is in this sense ``query-blind''. 
The exception is the recently published QAFD-RAG~\cite{zhou2026qafd}, to our knowledge the first system 
in which the traversal itself becomes query-aware: each edge is re-weighted by a combined function 
of the pairwise similarity and the similarity to the query. 
However, its flow-diffusion solver requires loading the entire graph into Python data structures, 
and edge-level re-weighting is awkward to express in the query language of a graph database.

Our method retains the core idea of QAFD-RAG --- that the traversal must account for the query. 
At the same time, we dispense with holding the graph in memory and with the combined edge re-weighting. 
The proposed algorithm spreads activation from seed entities extracted from the question 
across the knowledge graph. 
Each propagation step passes through a semantic gate whose intensity is set by the cosine similarity 
between the candidate entity's description and the question. 
The entire retrieval procedure --- seed-node mapping, three propagation iterations, 
selection of $K$ chains and entities, and context return --- is expressed as a single Cypher query 
executed in one round-trip to Neo4j, without moving any graph data outside the database. 
Experiments on two multi-hop benchmarks reveal three key findings.

\begin{enumerate}
\item \textbf{Query-aware propagation simultaneously improves accuracy and reduces retrieval latency.} 
On both datasets --- MuSiQue and 2WikiMultiHopQA --- replacing structural spreading activation 
with the proposed query-aware variant improves $F_1$ (+7.4 on MuSiQue, +3.6 on 2wiki) 
and reduces retrieval latency by a factor of 1.5 to 4.9. 
The improvement arises because the semantic gate prunes low-information branches at each iteration, 
thereby shrinking the working set on which the next iteration operates.

\item \textbf{Retrieval is a single Cypher query and loads no graph into memory.} 
Both HippoRAG and QAFD-RAG materialize the graph into Python data structures to perform the traversal. 
In contrast, the proposed method delegates the entire traversal to Neo4j's query planner; 
this makes the system trivial to deploy in production, parallelizes propagation along the activation 
front inside the database, and ensures concurrency safety.

\item \textbf{A simpler gate attains EM on par with the best query-aware system.} 
On MuSiQue the proposed method virtually matches QAFD-RAG on EM (32.80 vs 33.50) 
and outperforms the strongest of the purely structural baselines --- HippoRAG --- 
by 3.4 $F_1$ and 5.3 EM; on 2WikiMultiHopQA the PageRank-based methods retain the advantage.
\end{enumerate}

The full source code, the indexing procedure, and per-question evaluation logs are available at 
\url{https://github.com/kinfi4/qasa-graph-rag} to support reproduction of the results 
and further research.

\section{Related Work}
\label{sec:related}

The literature is organized by the degree to which the question influences retrieval. 
Methods that use the question only to select seed nodes are reviewed in 
Section~\ref{sec:related-structural}, and those that additionally let the question shape the 
graph traversal in Section~\ref{sec:related-queryaware}. 
Table~\ref{tab:comparison} presents this comparison at the level of architectural choices, 
indicating where the proposed method fits.

\subsection{Graph RAG with Purely Structural Traversal}
\label{sec:related-structural}

\textbf{GraphRAG}~\cite{edge2024graphrag} extracts entities and relations using a language model, 
applies the Leiden community-detection algorithm to the resulting graph, 
and builds a hierarchy of textual summaries; at query time the system answers either by 
aggregating community summaries (the ``global'' mode) or by selecting the most relevant 
entities (the ``local'' mode). 
This method is highly effective for query-focused summarization but is not designed to 
answer multi-hop factual questions.

\textbf{LightRAG}~\cite{guo2024lightrag} jointly indexes entities and relations and performs a 
dual-level retrieval that combines vector similarity over entity descriptions with keyword 
retrieval over the relation topics generated by the language model. 
Traversal here is implicit: each retrieved entity returns its one-hop neighborhood. 
LightRAG's main weakness on multi-hop benchmarks is the absence of explicit multi-hop expansion: 
relevant evidence often lies two or three hops from any seed node.

\textbf{HippoRAG}~\cite{gutierrez2024hipporag} casts retrieval as personalized 
PageRank~\cite{jeh2003ppr}, starting from a small set of phrase entities extracted from the 
question and matched to knowledge-graph nodes by their vector representations. 
The graph is built from OpenIE triples, so its nodes are phrase entities rather than passages. 
Each passage's score is computed at the end by aggregating the PageRank scores of the entities 
it mentions. 
\textbf{HippoRAG~2}~\cite{gutierrez2025hipporag2} refines the architecture: passage nodes are 
added to the phrase nodes, the seed-selection stage uses query-to-triple matching, 
and a ``recognition memory'' filter based on a language model is applied before traversal; 
PPR remains the traversal primitive.

\textbf{PathRAG}~\cite{chen2025pathrag} selects nodes by vector similarity to the query keywords, 
then computes paths between every pair of retrieved nodes on a network-flow principle: 
each path is scored by the amount of resource remaining after it is traversed with a fixed decay. 
The result is a set of relational paths presented to the language model in ascending order of 
reliability. 
This ordering mitigates the ``lost-in-the-middle'' positional bias~\cite{liu2024lost} 
characteristic of language models. 
As in the preceding systems, path scoring depends only on the structure.

\subsection{Query-Aware Graph Traversal}
\label{sec:related-queryaware}

\textbf{QAFD-RAG}~\cite{zhou2026qafd} is the closest prior work to ours in concept. 
There, retrieval is formulated as a flow-diffusion problem on an entity graph, in which each edge 
$(u,v)$ is re-weighted by the combined function 
$H_{\mathrm{sim}}(u,v) \cdot \bigl(a + b\,(H_{\mathrm{sim}}(u,q) + H_{\mathrm{sim}}(v,q))\bigr)$, 
where $H_{\mathrm{sim}}$ is a similarity score, $q$ is the query, and $a, b$ are hyperparameters 
(the authors choose $a = 1$, $b = 1/4$ in their experiments). 
The diffusion is solved by a primal-dual iterative procedure --- randomized coordinate descent 
under an $\varepsilon$-convergence criterion --- for which the authors prove subgraph-recovery guarantees. 
In practice, the number of iterations depends on the specific query, the seed entities, and the 
graph structure, and is not fixed in advance; combined with the per-edge flow state, this makes the 
algorithm difficult to express in the query language of a graph database, so the published 
implementation loads the graph into Python data structures. 
The proposed method builds on the same central idea --- that traversal must account for the query --- 
but replaces the combined edge re-weighting and the flow-diffusion solver with a single per-step gate 
that depends only on the target node and is expressible in the query language of a graph database.

\begin{table*}[t]
\centering
\caption{The proposed method among representative Graph RAG systems.}
\label{tab:comparison}
\footnotesize
\begin{tabular}{@{}p{0.14\textwidth} p{0.295\textwidth} p{0.215\textwidth} >{\centering\arraybackslash}p{0.085\textwidth} >{\centering\arraybackslash}p{0.105\textwidth}@{}}
\toprule
\textbf{Method} & \textbf{Traversal primitive} & \textbf{Query influence on traversal} & \textbf{Graph in memory} & \textbf{Retrieval execution} \\
\midrule
GraphRAG~\cite{edge2024graphrag} & Community detection + hierarchical summarization & seed selection only & yes & Python \\
LightRAG~\cite{guo2024lightrag} & One-hop neighborhood aggregation & seed selection only & yes & Python \\
HippoRAG~\cite{gutierrez2024hipporag} & Personalized PageRank from seed entities & seed selection only & yes & Python \\
HippoRAG~2~\cite{gutierrez2025hipporag2} & Personalized PageRank; phrase nodes + passage nodes & selection + triple filter & yes & Python \\
PathRAG~\cite{chen2025pathrag} & Flow-based path pruning & seed selection only & yes & Python \\
QAFD-RAG~\cite{zhou2026qafd} & Flow diffusion with combined edge re-weighting & per-step, on the edge (combined formula) & yes & Python \\
\textbf{Proposed method} & Spreading activation with a per-step semantic gate & per-step, on the node (cosine gate) & \textbf{no} & \textbf{Cypher} \\
\bottomrule
\end{tabular}
\end{table*}

\section{Methodology}
\label{sec:method}

\begin{figure*}[t]
\centering
\includegraphics[width=\textwidth]{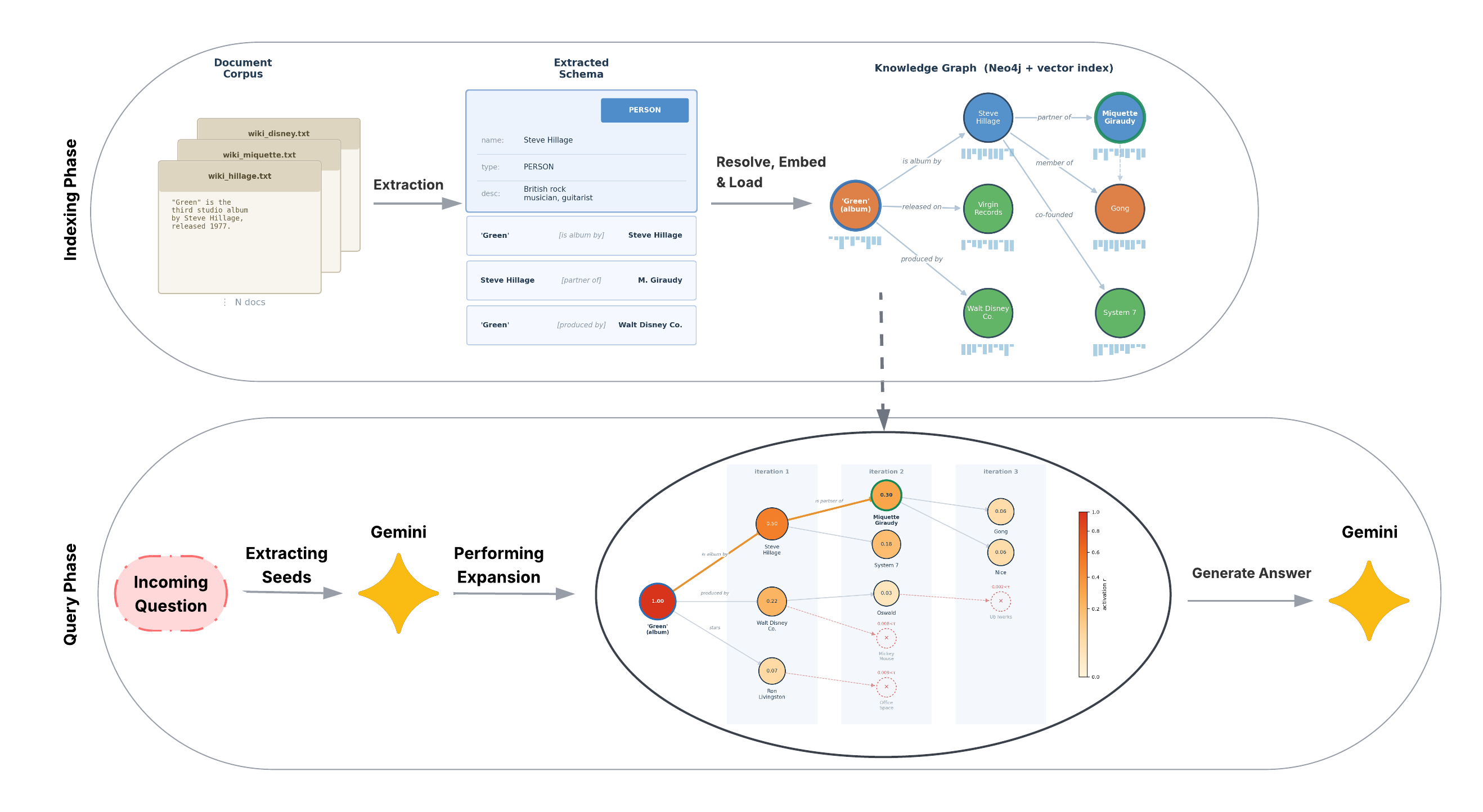}
\caption{Two-phase architecture.
The \textit{indexing phase} (top) extracts entities and relations from the corpus via an LLM,
resolves and embeds them, and stores the result as a typed knowledge graph in Neo4j.
The \textit{query phase} (bottom) uses an LLM to map the question to seed nodes,
runs spreading activation for $T$ iterations in a single Cypher query,
and passes the activated subgraph to an LLM to generate the answer.}
\label{fig:pipeline}
\end{figure*}

The system operates in two phases: the \textit{indexing phase} builds a typed Neo4j knowledge 
graph from the corpus (Section~\ref{sec:method-kg}), and the \textit{query phase} assembles the 
context for the language model in a single database round-trip 
(Section~\ref{sec:method-retrieval}). 
The overall architecture is shown in Fig.~\ref{fig:pipeline}.

\subsection{Knowledge Graph Construction}
\label{sec:method-kg}

\textbf{Entity and relation extraction.} 
Paragraphs that share a common title are merged into a single document, which is passed to the 
language model (Gemini~2.5 Flash) in a single call with a structured-output schema. 
The model returns a list of named entities, each with an assigned type from 
\{PERSON, ORGANIZATION, LOCATION, EVENT, WORK\} and a textual description, together with a list of 
binary relations --- triples \texttt{(source\_name, relation, target\_name)}. 
Extraction follows strict rules: no pronouns, full proper names only, and a mandatory description. 
Results are cached on disk, keyed by the document hash, so that repeated runs do not consume tokens.

\textbf{Entity resolution.} 
A second pass merges mentions of the same entity that appear in the corpus under different surface 
forms (for example, ``Louis~XIII'' and ``Louis~XIII of France''). 
Within each entity type, a set of merge candidates is built by a deterministic criterion: 
a pair of names forms an edge if the tokens of one are a subset of the tokens of the other; 
the connected components of this relation are grouped by a union-find structure. 
This pre-filtering avoids a quadratic number of language-model calls and narrows the model's task 
to verifying correctness within the already-formed clusters. 
Each non-empty cluster, with the candidate descriptions, is passed to the language model in a single 
call with a structured-output schema. 
The response contains zero or more disjoint groups; each group's canonical name is restricted to the 
set of input names, and an input entity may remain outside all groups, so within a single input 
cluster the model can distinguish several distinct canonical entities. 
The entity descriptions within each group are aggregated into a single description. 
After this, every alias mention in the triples is rewritten to its corresponding canonical name 
before being written to Neo4j; if rewriting collapses both endpoints of a triple to the same entity, 
that triple is discarded as an uninformative self-loop.

\begin{figure*}[t]
\centering
\includegraphics[width=\textwidth]{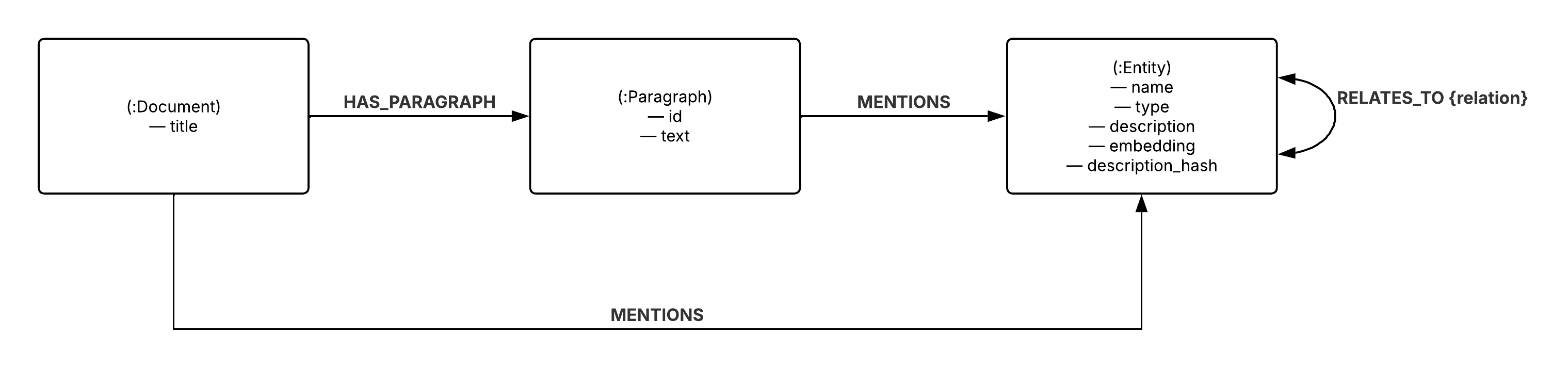}
\caption{Knowledge-graph schema. Document and paragraph nodes preserve the original structure of the 
corpus; \texttt{:MENTIONS} edges record entity mentions at two levels: the document (aggregated) and 
the paragraph (with exact location). Entities and their typed binary relations form the basis for 
multi-hop reasoning. The self-loop denotes the \texttt{:RELATES\_TO} edge, which connects two entities 
and stores the relation label as a property.}
\label{fig:schema}
\end{figure*}

\textbf{Writing to Neo4j.} 
The resolved entities and relations are written into the schema shown in Fig.~\ref{fig:schema}. 
Each \texttt{:Entity} node carries a name, a type, a textual description, a 768-dimensional vector 
representation of that description (computed with Google text-embedding-004), and a hash that ensures 
idempotency. 
A vector index is built over the embedding property using the Neo4j Graph Data Science library.

\subsection{Query-Aware Retrieval}
\label{sec:method-retrieval}

\textbf{Question entity extraction and seed-node mapping.} 
For each question, a structured language-model call extracts the named entities it mentions. 
Each question entity is mapped to graph nodes by an exact name match, or otherwise by a vector search 
in the entity-description index using its embedding. 
All such matches form the set $S$ of seed nodes. 
Each node $u \in S$ is assigned a matching score $s_u$ equal to~1 for exact name matches and to the 
cosine similarity between the question and the entity description for vector matches. 
Each $v \in S$ is given an initial activation $r^{(0)}(v) = s_v / \max_{u \in S} s_u$; 
for all other nodes $r^{(0)} = 0$. 
This normalization ensures that the best-matched seed node starts with an activation of~1, 
and that the activation threshold $\tau$ introduced below operates within the same absolute range 
regardless of the number and quality of the seed nodes.

\textbf{Spreading activation with query-aware weighting.} 
The traversal primitive is spreading activation, a classic mechanism in information 
retrieval~\cite{crestani1997spreading}. 
Let $\mathbf{e}_q$ denote the vector representation of the question and $\mathbf{e}_v$ the vector 
representation of an entity description. 
Propagation runs for a fixed number of iterations $T$ (by default $T = 3$). 
At iteration $t = 1, \dots, T$, each node $v$ receives an activation increment that combines the 
query-aware semantic gate $\sigma(v) = \max(\cos(\mathbf{e}_v, \mathbf{e}_q), 0)$ with a sum over the 
active front:
\begin{equation}
\begin{aligned}
\Delta r^{(t)}(v) &= \alpha \, \sigma(v) \sum_{u \in A^{(t)}(v)} r^{(t-1)}(u), \\
A^{(t)}(v) &= \bigl\{ u : (u,v) \in E,\; r^{(t-1)}(u) > \tau,\; \\
&\qquad\quad\; r^{(t-1)}(u) > r^{(t-1)}(v) \bigr\}.
\end{aligned}
\label{eq:delta}
\end{equation}

Here, $\alpha \in (0,1]$ is a fixed decay factor and $\tau$ a fixed threshold; the monotonic 
activation-decrease condition $r^{(t-1)}(u) > r^{(t-1)}(v)$ admits contributions only from nodes 
with higher activation, which prevents cyclic reinforcement. 
The threshold $\tau$ and the gate $\sigma$ play complementary roles in constraining the front: 
$\sigma$ prunes neighbors by their relevance to the query, $\tau$ by the absolute magnitude of the 
increment. 
Without the threshold, small increments would accumulate over $T$ iterations, propagation would cover 
the entire connected subgraph, and top-$K$ selection would gather many weakly activated entities. 
The activation update is therefore applied only when $\Delta r^{(t)}(v) > \tau$:
\begin{equation}
r^{(t)}(v) =
\begin{cases}
r^{(t-1)}(v) + \Delta r^{(t)}(v), & \Delta r^{(t)}(v) > \tau, \\
r^{(t-1)}(v), & \text{otherwise}.
\end{cases}
\label{eq:update}
\end{equation}

Unlike QAFD-RAG's combined weighting 
$H_{\mathrm{sim}}(u,v)\bigl(a + b\,(H_{\mathrm{sim}}(u,q) + H_{\mathrm{sim}}(v,q))\bigr)$~\cite{zhou2026qafd}, 
the proposed factor depends only on the target node and the query and is expressed as a single cosine 
in Cypher with no per-edge state; the monotonicity condition replaces flow conservation, so 
propagation runs for a fixed number of passes without an iterative solver. 
The procedure is formalized in Algorithm~\ref{alg:spreading}.

\begin{algorithm*}[t]
\caption{Query-Aware Spreading Activation for Graph RAG}
\label{alg:spreading}
\begin{algorithmic}[1]
\Require question $q$; knowledge graph $G = (V, E)$; depth $T$, decay factor 
    $\alpha$, threshold $\tau$; $K_e$ (entities to return), $K_c$ (chains to return)
\Ensure generated answer to $q$
\State $\{Q_1, \ldots, Q_m\} \gets \textsc{ExtractEntities}(q)$ \Comment{LLM structured call}
\State $S \gets \bigcup_i \textsc{MapToGraph}(Q_i)$ \Comment{exact match $\to$ vector fallback}
\State \textbf{for all} $v \in S$: \ $r^{(0)}(v) \gets s_v \,/\, \max_{u \in S} s_u$
\State \textbf{for all} $v \notin S$: \ $r^{(0)}(v) \gets 0$
\State $\mathbf{e}_q \gets \textsc{Embed}(q)$
\For{$t = 1$ \textbf{to} $T$}
    \ForAll{$v \in V$ s.t. $A^{(t)}(v) \neq \emptyset$} \Comment{$A^{(t)}(v)$ as in Eq.~\eqref{eq:delta}}
        \State $\sigma(v) \gets \max\bigl(\cos(\mathbf{e}_v, \mathbf{e}_q),\, 0\bigr)$
        \State $\Delta r^{(t)}(v) \gets \alpha \cdot \sigma(v) \cdot \sum_{u \in A^{(t)}(v)} r^{(t-1)}(u)$ \Comment{Eq.~\eqref{eq:delta}}
        \If{$\Delta r^{(t)}(v) > \tau$}
            \State $r^{(t)}(v) \gets r^{(t-1)}(v) + \Delta r^{(t)}(v)$
        \EndIf
    \EndFor
\EndFor
\State $E_{\mathrm{top}} \gets$ top-$K_e$ entities ranked by $r^{(T)}$
\State $C_{\mathrm{top}} \gets$ top-$K_c$ chains $(v_0, \ldots, v_l)$
\Statex \quad with $v_0 \in S$, $l \in [1, 4]$, $r^{(T)}(v_i) > 0$; ranked by mean activation
\State \Return $\textsc{Generate}\bigl(q, \textsc{Format}(P_{\mathrm{top}}, E_{\mathrm{top}})\bigr)$
\end{algorithmic}
\end{algorithm*}

Although each \texttt{:RELATES\_TO} edge is stored in the graph as directed --- from the subject to 
the object of the triple (Fig.~\ref{fig:schema}) --- spreading activation treats it as undirected: 
if the triple (Shevchenko, \emph{wrote}, Kobzar) is present in the graph, activation from ``Kobzar'' 
must reach Shevchenko just as it does in the reverse direction. 
Direction is retained only when chains are serialized for the language model, 
where it is critical to the semantic interpretation of the triple.

\textbf{Chain and entity selection.} 
After propagation, the $K_e$ entities with the highest $r^{(T)}$ are selected together with their 
stored descriptions, along with $K_c$ chains between any seed node $s \in S$ and any active node --- 
of length at most four edges, passing only through nodes touched by the spreading algorithm, 
ranked by mean activation along the chain. 
The four-edge length limit covers the typical reasoning depth in the target benchmarks 
(2--4 hops for MuSiQue) while bounding the combinatorial cost of selection, which grows 
exponentially with path length; the visited-node filter discards chains that pass through 
structurally adjacent nodes the spreading algorithm did not touch; and ranking by mean rather than 
minimum or total activation favors chains in which all nodes along the path are well activated, 
not merely their endpoints. 
Chains are deduplicated: the reversed sequences $(v_0, \dots, v_l)$ and $(v_l, \dots, v_0)$ are 
treated as a single chain, and a chain that is a prefix of a longer one is absorbed into it. 
Chains are linearized as a sequence of node--relation--node triples that preserve the relation 
direction, and entities as ``name (TYPE): description''.

\textbf{Context assembly and generation.} 
The context for the language model is formed by concatenating the linearized chains and the entity 
descriptions. 
The language model, driven by a structured-output schema, returns two fields, \texttt{answer} and 
\texttt{reasoning}; evaluation relies on the \texttt{answer} field alone.

\textbf{Parameter values.} 
Unless stated otherwise, all experiments use the following fixed values: activation threshold 
$\tau = 0.01$, selection limits $K_e = K_c = 30$, and a total fallback vector-search budget of 
5 candidates per question, distributed as $k_{\text{per\_entity}} = \max(1, \lfloor 5 / |S| \rfloor)$ 
across the entities extracted from the question. 
The propagation depth $T$ and the decay factor $\alpha$ are varied in the ablation study 
(Section~\ref{sec:ablation}); the default configuration is $T = 3$, $\alpha = 0.7$.

\subsection{Cypher Implementation}
\label{sec:method-cypher}
The entire retrieval procedure --- from seed-node mapping to returning the top-$K$ context --- 
is expressed as a single Cypher query that the client sends to Neo4j in one round-trip. 
The propagation state lives in a local \texttt{resource\_map} variable (a Cypher MAP) that is passed 
between phases through \texttt{WITH} clauses; no intermediate data is exchanged between the client 
and the database. 
The query is assembled from three sequential phases --- seed-node initialization, propagation 
iteration (repeated $T$ times), and chain-and-entity selection; the full text is given in 
the Appendix, and below we walk through the key Cypher constructs of each phase.

\textbf{Seed-node initialization.} 
For each entity extracted from the question (\texttt{UNWIND \$entity\_data AS ent}), an 
\texttt{OPTIONAL MATCH (exact:Entity \{name: ent.name\})} is performed first. 
The subsequent subquery \texttt{CALL (ent, exact) \{ ... UNION ALL ... \}} contains two branches 
guarded by \texttt{CASE WHEN exact IS NOT NULL} and \texttt{CASE WHEN exact IS NULL}: the first 
returns the exact match with a score of 1, while the second --- only when no exact match exists --- 
calls \texttt{db.index.vector.queryNodes} over the entity-description index and returns the $k$ 
nearest candidates with their cosine scores. 
Because of these guards, exactly one of the two branches executes for a given question entity, 
which realizes the ``exact name match, otherwise vector search'' semantics of 
Section~\ref{sec:method-retrieval} at the Cypher level, with no additional client-side logic. 
Matches from different question entities that point to the same node are deduplicated via 
\texttt{max(seed\_score)}; all scores are then normalized by dividing by the maximum, after which 
\texttt{apoc.map.fromPairs} builds the initial \texttt{resource\_map} from the normalized activations.

\textbf{Propagation iteration.} 
A single propagation step (Listing~\ref{lst:propagation}) is a subquery 
\texttt{CALL (resource\_map) \{ ... \}} that reads the current state and returns \texttt{deltas} --- 
a list of target nodes with their activation increments. 
The pass \texttt{UNWIND keys(resource\_map)} traverses the active front; the filter 
\texttt{WHERE src\_res > \$threshold} restricts it to nodes with activation above $\tau$. 
The condition 
\texttt{(resource\_map[dst.name] IS NULL OR resource\_map[dst.name] < src\_res)} realizes the 
monotonic-decrease condition $r^{(t-1)}(u) > r^{(t-1)}(v)$ from the definition of $A^{(t)}(v)$ in 
Eq.~\eqref{eq:delta}. 
The semantic gate $\sigma(v) = \max\bigl(\cos(\mathbf{e}_v, \mathbf{e}_q),\, 0\bigr)$ is computed as 
\texttt{gds.similarity.cosine(\allowbreak dst.description\_embedding, \$qe)} followed by 
\texttt{CASE WHEN sim > 0 THEN sim ELSE 0 END}. 
The increments to a single target node from different active neighbors are aggregated with 
\texttt{sum}, which matches the sum over the set $A^{(t)}(v)$ in Eq.~\eqref{eq:delta}; the condition 
\texttt{WHERE incoming > \$threshold} realizes the update threshold from Eq.~\eqref{eq:update}. 
Outside the subquery, \texttt{apoc.map.merge(resource\_map, apoc.map.fromPairs(...))} adds the 
\texttt{deltas} to the current \texttt{resource\_map} values, forming the state for the next iteration.

\begin{lstlisting}[float=*, floatplacement=t, caption={One step of query-aware propagation in Cypher. The block is repeated 
$T$ times in the query text as the client assembles it; each repetition takes \texttt{resource\_map} 
from the output of the previous one through a \texttt{WITH} clause.}, label={lst:propagation}]
CALL (resource_map) {
    UNWIND keys(resource_map) AS src_name
    WITH src_name, resource_map[src_name] AS src_res, resource_map
    WHERE src_res > $threshold
    MATCH (src:Entity {name: src_name})-[:RELATES_TO]-(dst:Entity)
    WHERE dst.description_embedding IS NOT NULL
      AND (resource_map[dst.name] IS NULL OR resource_map[dst.name] < src_res)
    WITH dst, src_res,
         gds.similarity.cosine(dst.description_embedding, $qe) AS sim
    WITH dst.name AS dst_name,
         sum(src_res * $decay *
             CASE WHEN sim > 0 THEN sim ELSE 0 END) AS incoming
    WHERE incoming > $threshold
    RETURN collect({name: dst_name, incoming: incoming}) AS deltas
}
WITH seed_names,
     apoc.map.merge(
       resource_map,
       apoc.map.fromPairs(
         [d IN deltas | [d.name,
                         coalesce(resource_map[d.name], 0) + d.incoming]]
       )
     ) AS resource_map
\end{lstlisting}

\textbf{Chain and entity selection.} 
After $T$ iterations, two subqueries read the final \texttt{resource\_map}. 
The first is a variable-length path pattern 
\texttt{MATCH path = (seed:Entity)-[:RELATES\_TO*1..4]-\allowbreak(target:Entity)} that instructs the 
Neo4j planner to enumerate all paths of length one to four edges between a seed node and any active 
node. 
The restriction ``only through nodes touched by propagation'' from Section~\ref{sec:method-retrieval} 
is expressed by the quantifier 
\texttt{ALL(n IN nodes(path) WHERE resource\_map[n.name] IS NOT NULL)}, while an additional condition 
\texttt{ALL(i IN range(0, size(nodes(path))-2) WHERE NOT nodes(path)[i] IN nodes(path)[(i+1)..])} 
discards paths with repeated nodes. 
Ranking by mean activation along the chain is implemented as 
\texttt{reduce(w = 0.0, n IN nodes(path) | w + coalesce(resource\_map[n.name], 0)) / size(nodes(path))}, 
after which \texttt{ORDER BY ... DESC LIMIT \$top\_k\_paths} keeps the top-$K_c$ paths. 
The second subquery ranks the keys of \texttt{resource\_map} by activation and returns the top-$K_e$ 
entities with their types and descriptions. 
The deduplication of reversed and absorbed chains described in Section~\ref{sec:method-retrieval} 
is performed on the client side over the returned result.

An architectural consequence of this implementation is that the full graph never leaves Neo4j and 
the entire traversal is performed by the query planner --- unlike HippoRAG~\cite{gutierrez2024hipporag} 
and QAFD-RAG~\cite{zhou2026qafd}, which materialize the graph into Python data structures.

\section{Experiments}
\label{sec:experiments}

\subsection{Datasets and Evaluation Protocol}
\label{sec:datasets}
Evaluation is conducted on two public multi-hop question-answering benchmarks. 
\textbf{MuSiQue}~\cite{trivedi2022musique} is built by composing two to four single-hop questions 
and is explicitly constructed so that single-hop heuristics fail to produce the correct answer; 
it is one of the hardest benchmarks of its class. 
\textbf{2WikiMultiHopQA}~\cite{ho2020twowiki} draws on structured evidence from Wikipedia 
(tables, infoboxes) and covers two question types --- ``bridge'' and ``comparison''. 
In both benchmarks, each question comes with its own set of paragraphs, both supporting and 
distractor. 
The paragraphs from the 1{,}000 sampled questions are merged into a single corpus per benchmark, 
from which the knowledge graph is built; this protocol follows HippoRAG~\cite{gutierrez2024hipporag}.

Table~\ref{tab:kgstats} reports the number of documents, paragraphs, entities, mentions, and 
relations obtained during the indexing phase on each benchmark. 
The MuSiQue corpus is roughly 1.7--2.2 times larger than 2Wiki on every metric and has substantially 
more mentions per entity (1.74 vs.\ 1.43); yet the graph-level density (relations per entity) is 
nearly identical in the two corpora. 
This matters when interpreting retrieval latency and input-token cost in Section~\ref{sec:latency}.

\begin{table*}[t]
\centering
\caption{Knowledge-graph statistics for each benchmark.}
\label{tab:kgstats}
\small
\begin{tabular}{@{}l rrrrr@{}}
\toprule
\textbf{Dataset} & \textbf{Documents} & \textbf{Paragraphs} & \textbf{Entities} & 
\textbf{Entity mentions} & \textbf{Relations} \\
\midrule
2WikiMultiHopQA & 6{,}322 & 6{,}367 & 31{,}143 & 44{,}535 & 40{,}784 \\
MuSiQue & 10{,}785 & 12{,}493 & 55{,}823 & 97{,}283 & 76{,}107 \\
\bottomrule
\end{tabular}
\end{table*}

The key comparison with prior systems is reported on the standard sample of $n = 1{,}000$ questions 
per dataset, matching the protocol of HippoRAG and QAFD-RAG. 
The internal ablation runs --- 11 configurations per dataset --- are performed on the first 
$n = 500$ questions of each dataset to limit the total inference cost. 
On this sample, the 95\% confidence interval for the mean $F_1$ is approximately $0.040$ for MuSiQue 
and $0.041$ for 2Wiki (a normal approximation consistent with a nonparametric bootstrap over 
10{,}000 resamples); we account for this interval width when interpreting closely-spaced results.

\subsection{Evaluation Metrics}
\label{sec:metrics}
We evaluate with two metrics: token-level $F_1$ and exact match (EM). 
Before computing EM and $F_1$, predictions and references are lowercased, and articles, punctuation, 
and redundant whitespace are removed.

\subsection{Infrastructure}
\label{sec:infra}
The graph store is Neo4j 5.26 Community Edition (LTS) with the Graph Data Science and APOC plugins, 
running in a local Docker container (Docker Desktop) with 4~GB of heap and 2~GB of page cache. 
All latency measurements are taken on a single machine --- a MacBook Pro with an Apple M1~Pro chip 
(10 cores: 8 performance and 2 efficiency), 16~GB of RAM, and a built-in NVMe SSD. 
Entity extraction, question-entity extraction, and answer generation all use Google Gemini~2.5 Flash 
at temperature~0; vector representations are computed with the 768-dimensional Google 
text-embedding-004 model.

\section{Results and Discussion}
\label{sec:results}

\subsection{Comparison with Prior Work}
\label{sec:comparison}
Table~\ref{tab:results} compares the proposed method with four published Graph RAG systems on the 
standard samples of $n = 1{,}000$ questions from both benchmarks. 
The figures for the competing methods are the QAFD-RAG authors' own reproductions of the baselines 
under a single experimental protocol~\cite{zhou2026qafd}; the details of that protocol for multi-hop 
questions are not documented in~\cite{zhou2026qafd} (see Section~\ref{sec:limitations}, limitation~1). 
With this caveat in mind: on MuSiQue the proposed method is essentially tied with QAFD-RAG on EM 
(32.80 vs.\ 33.50), trailing by 6.3 $F_1$; among the purely structural baselines it surpasses 
HippoRAG~\cite{gutierrez2024hipporag} --- the strongest of them in our sample --- by 5.3 EM and 
3.4 $F_1$, and GraphRAG and LightRAG by a wide margin. 
On 2WikiMultiHopQA, HippoRAG and QAFD-RAG retain an advantage of about 15 $F_1$ (discussed in 
Section~\ref{sec:phrasenode}), while the proposed method still exceeds GraphRAG and LightRAG. 
In summary, a single per-step cosine gate, expressed as one Cypher query without loading the graph 
into memory, attains EM close to QAFD-RAG on the harder of the two benchmarks, with a substantially 
simpler implementation.

\begin{table}[t]
\centering
\caption{Headline results on the standard $n = 1{,}000$ samples.}
\label{tab:results}
\footnotesize
\setlength{\tabcolsep}{5pt}
\begin{tabular}{@{}l cccc@{}}
\toprule
& \multicolumn{2}{c}{\textbf{MuSiQue}} & \multicolumn{2}{c}{\textbf{2WikiMultiHopQA}} \\
\cmidrule(lr){2-3} \cmidrule(lr){4-5}
\textbf{Method} & \textbf{F1} & \textbf{EM} & \textbf{F1} & \textbf{EM} \\
\midrule
GraphRAG~\cite{edge2024graphrag} & 39.40 & 17.60 & 15.20 & 7.00 \\
LightRAG~\cite{guo2024lightrag} & 1.40 & 0.10 & 8.20 & 1.00 \\
HippoRAG~\cite{gutierrez2024hipporag} & 38.33 & 27.55 & \textbf{70.33} & \textbf{61.16} \\
QAFD-RAG~\cite{zhou2026qafd} & \textbf{47.99} & \textbf{33.50} & 69.41 & 59.50 \\
\textbf{Proposed method} & 41.68 & 32.80 & 55.00 & 47.80 \\
\bottomrule
\end{tabular}
\end{table}

\subsection{Simultaneous Accuracy and Speed Gains}
\label{sec:gains}
Adding the query-aware semantic gate $\sigma(v)$ in Eq.~\eqref{eq:delta} simultaneously improves 
answer quality and reduces retrieval latency relative to uniform propagation on both datasets. 
In the control configuration \texttt{no-query-aware}, the gate $\sigma$ is replaced by the constant 1 
while the rest of the algorithm is unchanged, which isolates its contribution. 
On the standard sample of $n = 1{,}000$ (Table~\ref{tab:gate}), query-aware propagation raises $F_1$ 
by 7.4 points on MuSiQue (41.68 vs.\ 34.27) and by 3.6 points on 2Wiki (55.00 vs.\ 51.44); the gain 
is consistent with the control subsample of $n = 500$, on which the full ablation is performed 
(Section~\ref{sec:ablation}). 
At the same time, retrieval latency drops by a factor of 1.5--4.9 (detailed measurements are reported 
in Section~\ref{sec:latency}). 
Thus, the gate improves both metrics at once --- it is both more accurate and faster than uniform 
propagation.

\begin{table*}[t]
\centering
\caption{Contribution of the query-aware gate to answer quality.}
\label{tab:gate}
\small
\begin{tabular}{@{}l c cc cc@{}}
\toprule
\textbf{Configuration} & \textbf{N} & \multicolumn{2}{c}{\textbf{MuSiQue}} & 
\multicolumn{2}{c}{\textbf{2WikiMultiHopQA}} \\
\cmidrule(lr){3-4} \cmidrule(lr){5-6}
& & \textbf{F1} & \textbf{EM} & \textbf{F1} & \textbf{EM} \\
\midrule
\texttt{query-aware} & 500 & 46.71 & 36.6 & 55.91 & 49.0 \\
\texttt{no-query-aware} & 500 & 38.19 & 28.2 & 51.76 & 45.0 \\
\texttt{query-aware} & 1{,}000 & 41.68 & 32.8 & 55.00 & 47.8 \\
\texttt{no-query-aware} & 1{,}000 & 34.27 & 26.1 & 51.44 & 44.3 \\
\bottomrule
\end{tabular}
\end{table*}

The mechanism behind this simultaneous quality and speed gain is transparent. 
Without the semantic gate, every neighbor of an active node receives a positive inflow at each 
iteration, and the activation front grows roughly geometrically until it covers a substantial 
fraction of the graph. 
With the semantic gate, neighbors whose descriptions are unrelated to the query receive an inflow 
multiplied by a near-zero or exactly zero gate $\sigma(v)$, fall below the threshold $\tau$, and are 
pruned from the front of the next iteration. 
The result is a much smaller working set and --- because low-information entities never enter the 
top-$K$ ranking at all --- a less noisy context for the language model.

\subsection{Component Ablation}
\label{sec:ablation}
To measure the contribution of the key components, eleven configurations are evaluated on each 
dataset:
\begin{itemize}
\item \texttt{naive-vector} --- flat vector search over entity descriptions; serves as a graph-free 
baseline configuration.
\item \texttt{no-query-aware} --- in Eq.~\eqref{eq:delta}, the semantic gate $\sigma(v)$ is replaced 
by the constant 1 (uniform propagation).
\item \textbf{Depth sweep} --- $T \in \{1, 2, 3, 4\}$ with $\alpha = 0.7$ fixed.
\item \textbf{Decay sweep} --- $\alpha \in \{0.3, 0.5, 0.7, 0.9, 1.0\}$ with $T = 3$ fixed.
\item \textbf{The default configuration} ($T = 3$, $\alpha = 0.7$) belongs to both sweeps.
\end{itemize}

\begin{figure*}[t]
\centering
\includegraphics[width=\textwidth]{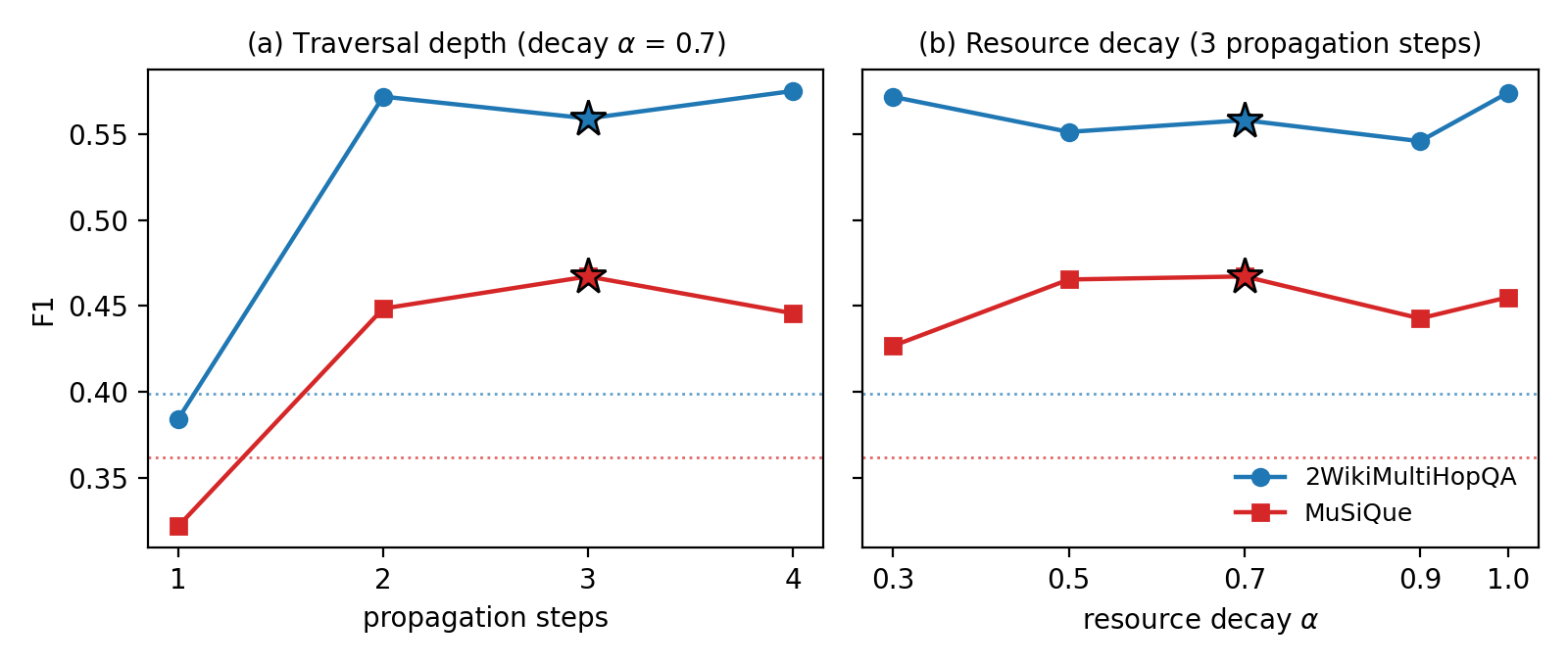}
\caption{Sensitivity analysis over the propagation depth (left) and the decay factor (right) at 
$n = 500$ per dataset. Stars mark the default configuration $T = 3$, $\alpha = 0.7$. The dashed 
horizontal lines show the \texttt{naive-vector} baseline $F_1$ on each dataset.}
\label{fig:ablation}
\end{figure*}

Fig.~\ref{fig:ablation} shows $F_1$ as a function of the propagation depth $T$ and the decay factor 
$\alpha$ on both datasets. 
Two patterns emerge. 
First, the transition from one to two steps ($T = 1 \rightarrow T = 2$) is the most impactful 
architectural decision in the entire system: it adds 18.8 $F_1$ on 2Wiki and 12.7 $F_1$ on MuSiQue 
--- more than any other change in Table~\ref{tab:ablation}. 
After two steps the curve flattens; on 2Wiki the values $T = 2, 3, 4$ are statistically 
indistinguishable, while on MuSiQue $T = 3$ is a local maximum, with $T = 4$ slightly degrading the 
result --- consistent with deeper expansion drawing irrelevant entities into the top-$K$ ranking. 
Second, the decay parameter is far less sensitive than the depth: the entire sweep 
$\alpha \in [0.3, 1.0]$ changes $F_1$ by only 2.8 points on 2Wiki and 4.0 points on MuSiQue, which 
in both cases lies within the confidence interval mentioned above. 
The default configuration $T = 3$, $\alpha = 0.7$ sits close to the maximum on both datasets, and the 
empirical surface is flat enough to avoid per-dataset hyperparameter tuning. 
All 11 configurations are listed in Table~\ref{tab:ablation}.

\begin{table*}[t]
\centering
\caption{Component ablation over eleven configurations ($n = 500$).}
\label{tab:ablation}
\footnotesize
\begin{tabular}{@{}l cccc@{}}
\toprule
& \multicolumn{2}{c}{\textbf{2WikiMultiHopQA}} & \multicolumn{2}{c}{\textbf{MuSiQue}} \\
\cmidrule(lr){2-3} \cmidrule(lr){4-5}
\textbf{Configuration} & \textbf{F1} & \textbf{EM} & \textbf{F1} & \textbf{EM} \\
\midrule
\texttt{naive-vector} (no graph) & 39.90 & 35.0 & 36.16 & 28.0 \\
\texttt{no-query-aware} (uniform propagation) & 51.76 & 45.0 & 38.19 & 28.2 \\
\texttt{steps-1} ($T = 1$, $\alpha = 0.7$) & 38.42 & 33.0 & 32.18 & 22.4 \\
\texttt{steps-2} ($T = 2$, $\alpha = 0.7$) & 57.18 & 50.6 & 44.84 & 34.4 \\
\textbf{\texttt{default} ($T = 3$, $\alpha = 0.7$)} & \textbf{55.91} & \textbf{49.0} & \textbf{46.71} & \textbf{36.6} \\
\texttt{steps-4} ($T = 4$, $\alpha = 0.7$) & 57.51 & 50.6 & 44.57 & 34.3 \\
\texttt{decay-0.3} ($T = 3$, $\alpha = 0.3$) & 57.17 & 50.2 & 42.68 & 31.4 \\
\texttt{decay-0.5} ($T = 3$, $\alpha = 0.5$) & 55.13 & 48.4 & 46.54 & 36.0 \\
\texttt{decay-0.9} ($T = 3$, $\alpha = 0.9$) & 54.59 & 47.6 & 44.27 & 34.6 \\
\texttt{decay-1.0} ($T = 3$, $\alpha = 1.0$) & 57.40 & 50.4 & 45.50 & 36.4 \\
\bottomrule
\end{tabular}
\end{table*}

\subsection{Latency and Input-Token Cost}
\label{sec:latency}
Table~\ref{tab:latency} summarizes retrieval latency, generation latency, and input-token cost per 
query for the five configurations that span the full propagation-depth ladder. 
Retrieval latency increases monotonically with $T$ on both corpora: from 0.01~s ($T = 1$) to 0.30~s 
($T = 4$) on 2Wiki, and from 0.03~s to 3.51~s on MuSiQue. 
At the same depth $T = 3$, uniform propagation (\texttt{no-query-aware}) is 4.9 times slower than the 
gated configuration on 2Wiki (0.44~s vs.\ 0.09~s) and 1.5 times slower on MuSiQue 
(2.16~s vs.\ 1.45~s), confirming that semantic pruning compresses the working set at each iteration. 
Increasing the number of iterations to four doubles the latency without any $F_1$ gain 
(Table~\ref{tab:ablation}), confirming saturation at $T = 3$. 
The input-token cost is substantially higher on the larger corpus: the default configuration uses 
14.1k input tokens per query for MuSiQue versus 4.5k for 2Wiki, consistent with the higher number of 
mentions per entity in MuSiQue (1.74 vs.\ 1.43) and, accordingly, with the longer synthesized 
descriptions per node.

\begin{table*}[t]
\centering
\caption{Latency and input-token cost per query ($n = 500$).}
\label{tab:latency}
\footnotesize
\begin{tabular}{@{}l ccc ccc@{}}
\toprule
& \multicolumn{3}{c}{\textbf{2WikiMultiHopQA}} & \multicolumn{3}{c}{\textbf{MuSiQue}} \\
\cmidrule(lr){2-4} \cmidrule(lr){5-7}
\textbf{Configuration} & \textbf{Retrieval (s)} & \textbf{Generation (s)} & \textbf{Input tokens} & 
\textbf{Retrieval (s)} & \textbf{Generation (s)} & \textbf{Input tokens} \\
\midrule
\texttt{no-query-aware} (uniform) & 0.44 & 0.90 & 5{,}137 & 2.16 & 1.43 & 17{,}976 \\
\texttt{steps-1} ($T = 1$, $\alpha = 0.7$) & 0.01 & 0.84 & 1{,}605 & 0.03 & 0.89 & 3{,}801 \\
\texttt{steps-2} ($T = 2$, $\alpha = 0.7$) & 0.02 & 0.88 & 3{,}617 & 0.59 & 1.08 & 9{,}494 \\
\textbf{\texttt{default} ($T = 3$, $\alpha = 0.7$)} & \textbf{0.09} & \textbf{0.86} & \textbf{4{,}505} & \textbf{1.45} & \textbf{1.39} & \textbf{14{,}130} \\
\texttt{steps-4} ($T = 4$, $\alpha = 0.7$) & 0.30 & 0.89 & 4{,}749 & 3.51 & 1.46 & 16{,}727 \\
\bottomrule
\end{tabular}
\end{table*}

\subsection{Per-Benchmark Analysis}
\label{sec:phrasenode}
The results on the two benchmarks differ substantially. 
On MuSiQue, where each question is composed of two to four single-hop questions over heterogeneous 
Wikipedia material, the proposed method is essentially tied with QAFD-RAG on EM and exceeds HippoRAG; 
on this benchmark the per-step semantic gate proves sufficient to locate and traverse the relevant 
reasoning chain without the additional combined-weighting mechanism of QAFD-RAG. 
On 2WikiMultiHopQA, by contrast, both HippoRAG and QAFD-RAG retain an advantage of about 15 $F_1$. 
We attribute this gap to two structural properties of the benchmark to which the OpenIE-based 
phrase-node architectures (HippoRAG, QAFD-RAG) are better suited.

First, \textbf{the granularity of entity representation}. 
Our offline phase extracts entities under a fixed type schema of five classes 
\{PERSON, ORGANIZATION, LOCATION, EVENT, WORK\} (Section~\ref{sec:method-kg}), whereas a noticeable 
fraction of the gold answers in 2Wiki are attributes that belong to none of these classes: dates, 
nationalities, languages, professions, and yes/no answers. 
Such facts do not exist as separate nodes in our graph --- they are only mentioned, in aggregate, 
within the textual description of their bearer entity, so spreading activation cannot reach them 
individually and they do not enter the top-$K$ directly. 
HippoRAG builds its graph from OpenIE triples without a fixed type schema, so phrase nodes such as 
``1845'', ``British'', or ``French painter'' are directly accessible. 
This gap is amplified by the fact that 2Wiki is built on structured fragments of Wikipedia 
(tables, infoboxes) that OpenIE decomposes especially cleanly into atomic triples.

Second, \textbf{the nature of ``comparison'' questions}, which make up a noticeable fraction of 
2Wiki. 
Questions such as ``Who was born earlier --- A or B?'' or ``Are both A and B located in country C?'' 
do not reduce to traversing a single reasoning chain: one must find two specific facts about 
already-given entities in parallel and compare them at the language-model level. 
Graph traversal adds little value here, since the key operation is not finding a connection between 
entities but selecting the right attributes. 
The proposed method is optimized precisely for the chain-inference questions that dominate MuSiQue.

The proposed method is not adapted to either of these regimes. 
A direction for removing these limitations --- extending the graph schema with heterogeneous nodes 
(paragraphs, phrase nodes), as is done in HippoRAG~2~\cite{gutierrez2025hipporag2} --- is discussed 
in Section~\ref{sec:future}.

\section{Limitations}
\label{sec:limitations}
We note the following limitations.
\begin{enumerate}
\item \textit{Mismatch in comparison conditions.} 
The figures for the competing methods in Table~\ref{tab:results} are the QAFD-RAG authors' own 
reproductions of the baselines under a single experimental protocol~\cite{zhou2026qafd}. 
The composition of the evaluation sample on which these figures were obtained in~\cite{zhou2026qafd} 
is not provided, so its overlap with our sample of $n = 1{,}000$ cannot be established. 
In~\cite{zhou2026qafd} the main language model is reported as GPT-4o-mini, whereas the proposed 
system uses Gemini 2.5 Flash. 
Both models belong to the same price tier and the same functional class of lightweight (Flash/mini) 
models from leading providers, so a substantial advantage for either generator is unlikely. 
The API budget allocated to this work permitted the use of Gemini only, so a full reproduction of 
the four competing systems with the same model for a fully controlled comparison was beyond that 
budget; the comparison in Table~\ref{tab:results} is made under the assumption that the experimental 
protocols are close.

\item \textit{The gap on 2WikiMultiHopQA.} 
The proposed method trails HippoRAG and QAFD-RAG by about 15 $F_1$ on 2Wiki. 
A structural explanation of this gap is given in Section~\ref{sec:phrasenode}; architectural 
extensions that could narrow it are discussed in Section~\ref{sec:future}.

\item \textit{Sample size for the control experiments.} 
The control runs are performed on $n = 500$ questions to limit the total inference cost; the 
corresponding 95\% confidence interval (Section~\ref{sec:datasets}) should be taken into account 
when interpreting these experiments. 
The key comparison in Table~\ref{tab:results} is performed on the standard samples of $n = 1{,}000$.

\item \textit{Unmeasured sensitivity to the language model.} 
All language-model calls in the proposed system --- entity extraction during indexing, 
question-entity extraction, and answer generation --- use the same Gemini 2.5 Flash model. 
The system's sensitivity to a different extractor or generator has not been evaluated. 
In particular, the discriminative power of the gate $\sigma(v)$ depends on how strongly the 
language-model-synthesized entity description focuses on the entity's identifying rather than 
contextual features; this interaction with the properties of the extractor remains unmeasured.
\end{enumerate}

\section{Conclusion}
\label{sec:conclusion}
We have proposed a query-aware spreading-activation method for Graph RAG, implemented as a single 
Cypher query to Neo4j. 
The algorithm propagates activation from the seed entities extracted from the question across the 
knowledge graph for a fixed number of iterations: at each step, neighbor contributions are 
multiplied by a per-step cosine gate that depends only on the current node and the query, and a 
monotonicity condition restricts propagation to nodes with higher activation. 
Unlike HippoRAG~\cite{gutierrez2024hipporag} and QAFD-RAG~\cite{zhou2026qafd}, which load the full 
graph into Python data structures (and QAFD-RAG additionally solves a flow-diffusion problem 
iteratively to a convergence criterion, so that the number of solver steps varies from query to 
query and cannot be expressed as a fixed Cypher query), the entire retrieval procedure --- from 
seed-node mapping to the selection of the top-$K$ chains and entities --- is expressed in a single 
Cypher query and executed directly by the Neo4j query planner. 
As a consequence, the size of the knowledge graph is not bounded by the amount of RAM, deploying the 
system reduces to running Neo4j, and indexing, data integrity, backup, and replication are provided 
by the database's built-in mechanisms.

On MuSiQue~\cite{trivedi2022musique} the method is essentially tied with 
QAFD-RAG~\cite{zhou2026qafd} on exact match and exceeds HippoRAG~\cite{gutierrez2024hipporag} by 
3.4 $F_1$ and 5.3 EM. 
On 2WikiMultiHopQA~\cite{ho2020twowiki} the purely structural HippoRAG and the flow-diffusion 
QAFD-RAG retain an advantage of about 15 $F_1$. 
A controlled comparison with a uniform variant of the same scheme (\texttt{no-query-aware}, 
Tables~\ref{tab:gate} and~\ref{tab:latency}) shows that adding the query-aware gate on both 
benchmarks simultaneously improves accuracy and reduces retrieval latency by a factor of 1.5--4.9 
--- the gate prunes low-information branches at each iteration.

\label{sec:future}%
We highlight two promising directions. 
First, extending spreading activation to heterogeneous nodes --- in particular, to the paragraph 
nodes already present in the schema, as is done in HippoRAG~2~\cite{gutierrez2025hipporag2} --- could 
close part of the gap on 2Wiki while preserving expressibility in Cypher. 
Second, the proposed gate can be embedded in an iterative retrieval scheme in which the results of 
one pass serve as the seed nodes of the next, which would help locate intermediate ``bridge'' 
entities without reloading the full graph at each iteration.

\section*{Declaration on the Use of AI Tools}
AI-based tools (in particular, Claude) were used to edit and linguistically improve the text of the 
manuscript and/or as an auxiliary aid while writing the code for the experiments and analysis, under 
the authors' supervision. 
All scientific decisions --- experiment design, data interpretation, and conclusions --- belong to 
the authors.


\onecolumn
\section*{Appendix: Full Text of the Cypher Query}
The full query, which implements the procedure of Section~\ref{sec:method-cypher}, is assembled by 
the client by concatenating three fragments: seed-node initialization, the propagation block 
(repeated $T$ times), and chain-and-entity selection. The query parameters are: 
\texttt{\$entity\_data} --- the list of entities extracted from the question, with fields 
\texttt{name} and \texttt{emb} (the embedding of the description); \texttt{\$qe} --- the embedding 
of the question; \texttt{\$decay} --- the decay factor $\alpha$; \texttt{\$threshold} --- the 
threshold $\tau$; \texttt{\$top\_k\_entities}, \texttt{\$top\_k\_paths} --- the limits $K_e$, $K_c$; 
\texttt{\$k\_per\_entity} --- the number of fallback vector-search candidates per question entity.

\begin{lstlisting}
// === Fragment 1: seed-node initialization ===
UNWIND $entity_data AS ent
OPTIONAL MATCH (exact:Entity {name: ent.name})

CALL (ent, exact) {
    UNWIND CASE WHEN exact IS NOT NULL THEN [exact] ELSE [] END AS e
    RETURN e.name AS seed_name, 1.0 AS seed_score
  UNION ALL
    UNWIND CASE WHEN exact IS NULL THEN [ent] ELSE [] END AS e
    CALL db.index.vector.queryNodes(
        'entity_description_embeddings', $k_per_entity, e.emb
    ) YIELD node, score
    RETURN node.name AS seed_name, score AS seed_score
}

WITH seed_name, max(seed_score) AS seed_score
WITH collect({name: seed_name, score: seed_score}) AS raw,
     max(seed_score) AS mx
WITH [s IN raw | s.name] AS seed_names,
     apoc.map.fromPairs([s IN raw | [s.name, s.score / mx]]) AS resource_map

// === Fragment 2: propagation block (repeated T times) ===
CALL (resource_map) {
    UNWIND keys(resource_map) AS src_name
    WITH src_name, resource_map[src_name] AS src_res, resource_map
    WHERE src_res > $threshold
    MATCH (src:Entity {name: src_name})-[:RELATES_TO]-(dst:Entity)
    WHERE dst.description_embedding IS NOT NULL
      AND (resource_map[dst.name] IS NULL OR resource_map[dst.name] < src_res)
    WITH dst, src_res,
         gds.similarity.cosine(dst.description_embedding, $qe) AS sim
    WITH dst.name AS dst_name,
         sum(src_res * $decay *
             CASE WHEN sim > 0 THEN sim ELSE 0 END) AS incoming
    WHERE incoming > $threshold
    RETURN collect({name: dst_name, incoming: incoming}) AS deltas
}
WITH seed_names,
     apoc.map.merge(
       resource_map,
       apoc.map.fromPairs(
         [d IN deltas | [d.name,
                         coalesce(resource_map[d.name], 0) + d.incoming]]
       )
     ) AS resource_map

// === Fragment 3: chain-and-entity selection ===
CALL (seed_names, resource_map) {
    MATCH path = (seed:Entity)-[:RELATES_TO*1..4]-(target:Entity)
    WHERE seed.name IN seed_names
      AND resource_map[target.name] IS NOT NULL
      AND seed <> target
      AND ALL(n IN nodes(path) WHERE resource_map[n.name] IS NOT NULL)
      AND ALL(i IN range(0, size(nodes(path))-2)
          WHERE NOT nodes(path)[i] IN nodes(path)[(i+1)..])
    WITH reduce(w = 0.0, n IN nodes(path) |
                  w + coalesce(resource_map[n.name], 0))
             / size(nodes(path)) AS path_weight,
         [n IN nodes(path) | n.name] AS names,
         [i IN range(0, size(relationships(path))-1) |
             {relation: relationships(path)[i].relation,
              forward: startNode(relationships(path)[i]) = nodes(path)[i]}
         ] AS edges
    ORDER BY path_weight DESC
    LIMIT $top_k_paths
    RETURN collect({names: names, edges: edges,
                    path_weight: path_weight}) AS paths
}

CALL (resource_map) {
    UNWIND keys(resource_map) AS name
    WITH name, resource_map[name] AS resource
    ORDER BY resource DESC LIMIT $top_k_entities
    MATCH (e:Entity {name: name})
    RETURN collect({
        name: e.name, resource: resource,
        type: e.type, description: e.description
    }) AS entities
}

RETURN seed_names, paths, entities
\end{lstlisting}

\end{document}